\pdfoutput=1
\documentclass[10pt,twocolumn,letterpaper]{article}

\usepackage[pagenumbers]{cvpr} % To force page numbers, e.g. for an arXiv version

\usepackage{graphicx}
\usepackage{amsmath}
\usepackage{amssymb}
\usepackage{booktabs}
\usepackage{multirow}
\usepackage{color}
\usepackage{colortbl}
\usepackage{makecell}
\usepackage{threeparttable}
\usepackage{graphicx}
\usepackage{subcaption}
\usepackage{bigstrut}
\usepackage{amsmath,amsfonts}
\usepackage{arydshln}  
\definecolor{mycolor}{rgb}{0.9, 0.9, 0.9} % 淡紫

\usepackage[pagebackref,breaklinks,colorlinks]{hyperref}

\usepackage[capitalize]{cleveref}
\crefname{section}{Sec.}{Secs.}
\Crefname{section}{Section}{Sections}
\Crefname{table}{Table}{Tables}
\crefname{table}{Tab.}{Tabs.}

\def\cvprPaperID{*****} % *** Enter the CVPR Paper ID here
\def\confName{CVPR}
\def\confYear{2022}

\begin{document}

%%%%%%%%% TITLE - PLEASE UPDATE
\title{Task-Balanced Distillation for Object Detection}

\author{Ruining Tang$^{1}$, Zhenyu Liu$^{1}$, Yangguang Li$^{2}$, Yiguo Song$^{1}$, Hui Liu$^{1}$, Qide Wang$^{1}$,\\ Jing Shao$^{2}$, Guifang Duan$^{1 \thanks{Corresponding author}}$, Jianrong Tan$^{1}$\\
$^{1}$ State Key Lab of CAD\&CG, Zhejiang University, $^{2}$ SenseTime Research\\
%the State Key Laboratory of CAD\&CG, Zhejiang University,
{\tt\small \{tangruining,liuzy,ygsong,liuhui2017,btwqd,gfduan,egi\}@zju.edu.cn}\\
{\tt\small liyangguang@sensetime.com, shaojing@senseauto.com}
}

% address
\maketitle
%%%%%%%%% ABSTRACT
\begin{abstract}
	Mainstream object detectors are commonly constituted of two sub-tasks, including classification and regression tasks, implemented by two parallel heads. This classic design paradigm inevitably leads to inconsistent spatial distributions between classification score and localization quality (IOU). Therefore, this paper alleviates this misalignment in the view of knowledge distillation. First, we observe that the massive teacher achieves a higher proportion of harmonious predictions than the lightweight student. Based on this intriguing observation, a novel Harmony Score (HS) is devised to estimate the alignment of classification and regression qualities. HS models the relationship between two sub-tasks and is seen as prior knowledge to promote harmonious predictions for the student. Second, this spatial misalignment will result in inharmonious region selection when distilling features. To alleviate this problem, a novel Task-decoupled Feature Distillation (TFD) is proposed by flexibly balancing the contributions of classification and regression tasks. Eventually, HD and TFD constitute the proposed method, named Task-Balanced Distillation (TBD). Extensive experiments demonstrate the considerable potential and generalization of the proposed method. Specifically, when equipped with TBD, RetinaNet with ResNet-50 achieves 41.0 mAP under the COCO benchmark, outperforming the recent FGD and FRS. 
\end{abstract}
%%%%%%%%% BODY TEXT
\section{Introduction}
\label{sec:intro}
In recent years, the development of object detectors has drawn wide attention of the computer vision community, especially with the growth of convolutional neural networks (CNNs). As a fundamental pillar of the computer vision task, object detectors have been universally involved in all walks of life, such as autonomous driving, security monitoring, and pedestrian detection. In general, mainstream object detectors \cite{ren2015faster, cai2018cascade, li2020generalized, lin2017focal, tian2019fcos, zhang2020bridging} can be approximately divided into two-stage detectors \cite{ren2015faster, cai2018cascade} and one-stage detectors \cite{li2020generalized, lin2017focal, tian2019fcos, zhang2020bridging} depending on whether the region proposal network (RPN) is implemented.

\begin{figure}[t]
	\centering
	\includegraphics[width=0.95\linewidth]{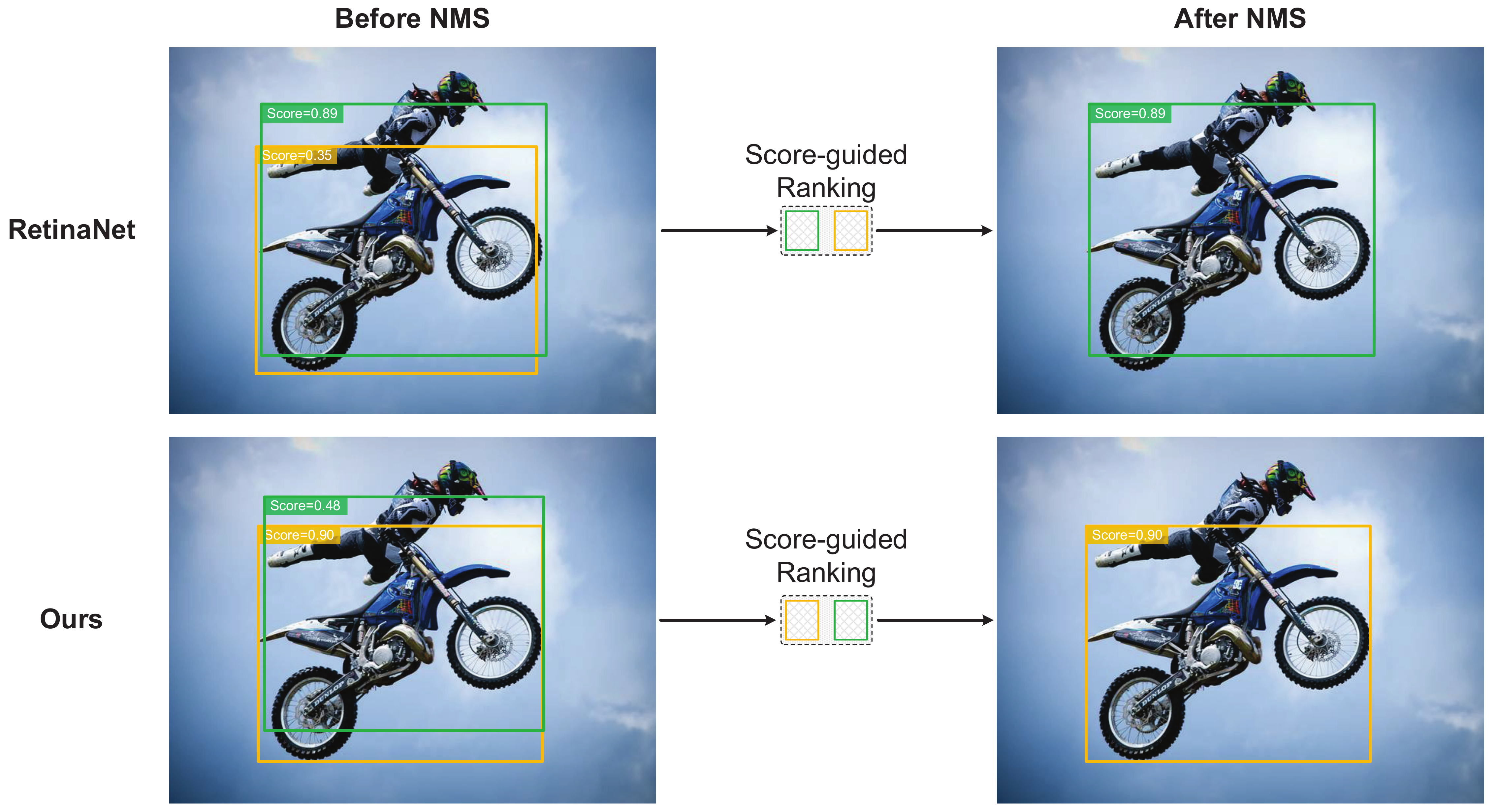}
	\caption{Visualization of the NMS mechanisms between the vanilla RetinaNet (top sub-figures) and the proposed model (bottom sub-figures). For ease of understanding, only two samples are shown here. The vanilla RetinaNet generates inconsistent predictions, leading to inaccurate preservation (green bbox). After equipping with the proposed method, the high-quality candidate (orange bbox) is conclusively preserved. }
	\label{fig1}
	%\Description{A woman and a girl in white dresses sit in an open car.}
\end{figure}

To generate both the location coordinates and the corresponding label for an object, modern object detectors typically adopt a multi-task pipeline, which consists of a classification branch and regression branch, implemented by two parallel heads. However, this parallel implementation may lead to inconsistent distributions of classification score and regression quality (IOU). As shown in the top sub-figures of Fig.\ref{fig1}, the vanilla RetinaNet outputs inconsistent predictions due to the overlap between the person and motorcycle. Specifically, the green candidate contains a high score but a low IOU, whereas the orange one has an accurate bbox but a low score. When the post-procedure (e.g., Non-Maximum Suppression) is executed, the green one with a larger score will be reserved since the classification score is used as a general criterion for NMS ranking. As a result, the prediction with an accurate bbox (orange one) may be mistakenly filtered. Generally speaking, this incorrect filtering can be attributed to the inconsistent distribution between classification and localization accuracy.

Previous works \cite{tian2019fcos, kong2020foveabox, jiang2018acquisition, wang2021reconcile, feng2021tood, gao2021decoupled} attempt to overcome this problem in three ways, including recomposing the NMS score via adding an additional head (\textit{i.e.}, IOUNet \cite{jiang2018acquisition}, Centerness \cite{tian2019fcos, zhang2020bridging}), focusing on consistent regions \cite{tian2019fcos, kong2020foveabox}, and enhancing the dependency between classification and regression tasks to output more harmonious predictions \cite{wang2021reconcile, feng2021tood}. Although these studies have made remarkable progress in alleviating the influence of the inconsistent spatial distributions, the motivations and solutions are derived from the detector itself. Different from these methods above, this paper alleviates this inherent problem in the view of knowledge distillation by designing a customized teacher-student training workflow.

To elicit the proposed method, we meticulously compare the behaviour between the teacher and student models in handling this inharmonious distribution. A valuable observation is derived as follows. \textbf{In general, the teacher is more inclined to generate harmonious predictions than the student.} We count the IOU distributions of easy-classified predictions, as demonstrated in Table \ref{tab:tab0}. Overall, the teacher model performs superiorly at achieving highly consistent predictions (\textit{i.e.}, 69.2 vs. 67.43), while maintaining fewer inharmonious predictions (\textit{i.e.}, 29.72 vs. 31.68). This observation indicates that owing to the disparate distributions of classification and regression qualities, some easy-classified samples may suffer from inaccurate locations for the student (\textit{e.g.}, the motorcycle appeared in Fig. \ref{fig1}). Therefore, one meaningful question is whether the student can generate more harmonious predictions with the assistance of the teacher model.

In addition, we observe that \textbf{the inharmonious distributions of two sub-tasks will affect the selection of substantial areas when distilling features.} For transferring the intermediate features from the teacher to the student, a meaningful route \cite{wang2019distilling, kang2021instance, zhixing2021distilling, li2021knowledge, zhang2020improve, dai2021general, yang2022focal} is how to screen the significant regions. Previous works \cite{zhixing2021distilling, li2021knowledge} attempt to generate spatial masks to denote the meaningful areas by using the predictions of the classification branch. However, considering the inharmonious distributions between classification and localization accuracy\cite{feng2021tood}, purely utilizing the classification information might result in sub-optimal region selection. Therefore, another critical question is whether the classification and localization information can be fully utilized to guide the feature imitation.
\begin{table}[t]\centering
	\caption{The IOU distributions of easy-classified predictions on COCO \textit{minival} split. Specifically, predictions with scores larger than 0.9 are counted. }
	\label{tab:tab0}
	\resizebox{0.47\textwidth}{!}{
	\begin{tabular}{lccc}
		\toprule
		Model & $IOU\ge 0.9$  & $0.5\le IOU <  0.9$ & $IOU <  0.5$  \\
		\midrule
		Teacher & 69.2\%& 29.72\% & 1.0\%  \\
		Student & 67.43\% & 31.68\% & 0.9\% \\
		\midrule
		HD (ours) & 70.97\%  & 28.4\%  & 0.58\%   \\
		Relative gains & \textbf{+5.25\%} & \textbf{-10.35\%}& \textbf{-35.56\%} \\
		\bottomrule
	\end{tabular}}
\end{table}

A novel \textbf{Harmony Distillation (HD)} component is devised to achieve the transformation of harmonious predictions. Firstly, the Harmony Score (HS) is defined to quantitatively describe the deviation of the classification score and the corresponding regression quality. In particular, a large HS implies the classification score is positively correlated with the regression quality and vice versa. Secondly, the HD is derived by aligning the HS between teacher and student models. The proposed HD affords prior knowledge that models the relationship between classification and regression to assist the generation of high-quality predictions for the student. As presented in Tab. \ref{tab:tab0}, the proportion of harmonious predictions is significantly improved, even surpassing that of the teacher. 

To achieve the effective feature imitation, a new \textbf{Task-decoupled Feature Distillation (TFD)} is devised to integrate the information from both classification and regression tasks. The classification-aware and localization-aware masks are firstly obtained by using the corresponding predictions. Furthermore, instead of combining these masks with a heuristic weight scheme, we propose a Task-collaborative Weight Generation (TWG) module to balance the contributions of classification and regression tasks. Concretely, TWG dynamically assigns the task-aware weights according to both teacher's and student's predictions.

The proposed \textbf{Task-Balanced Distillation (TBD)} consists of the above HD and TFD, jointly considering the properties of classification and localization. To evaluate the effectiveness of the proposed method, we conduct experiments on the Pascal VOC \cite{everingham2010pascal}, COCO \cite{lin2014microsoft}, TJU-DHD \cite{pang2020tju}, and Cityscapes \cite{cordts2016cityscapes} benchmarks. Abundant experimental results demonstrate the effectiveness and generalization of the proposed method. For instance, when equipped with the proposed TBD, RetinaNet-R50 achieves 41.0 mAP, surpassing the baseline by a large margin (\textit{i.e.}, 3.6 mAP), even outperforming the current SOTA methods such as FRS \cite{zhixing2021distilling} and FGD \cite{yang2022focal}. 

To sum up, the contributions of this paper are summarized as follows:
\begin{itemize}
	\item A new Harmony Score (HS) is firstly defined to capture the relationship between classification and regression qualities. Then a novel Harmony Distillation (HD) is proposed to assist the generation of harmonious predictions for the student.
	\item A novel Task-decoupled Feature Distillation (TFD) is devised to mimic the intermediate features. The classification and regression masks are synthetically combined by balancing the contributions of these two tasks.
	\item Abundant experiments among various datasets and detectors are conducted. In addition, we achieve effective distillation between homogeneous (CNN to CNN) and heterogeneous (Transformer to CNN) backbones. The proposed method is easily plugged in and achieves SOTA performance.
\end{itemize}

%-------------------------------------------------------------------------
\section{Related Works}
\subsection{Object Detection}
With the wide application of deep learning technology, the architecture of object detectors has progressively moved towards an end-to-end pipeline. Inherited from the ideology of pioneers, Faster RCNN \cite{ren2015faster} innovatively applies a two-stage scheme to detect objects. The two-stage detectors primarily consist of two core components. The first component completes the generation of potential candidates, whereas the second one further enables precise classification and regression based on these candidates. This architecture is popularized by its variants \cite{cai2018cascade, dai2016r, he2017mask, fan2022adaptive, peng2022context}. On the contrary, well-known one-stage detectors \cite{redmon2016you, liu2016ssd, lin2017focal, feng2021tood, tian2019fcos, li2020generalized, miao2022balanced, su2022dsla} vastly simplified this two-step paradigm, and they directly make predictions based on the learned features. 
\subsection{Harmonious Predictions}
The content of the inharmonious prediction is originally derived from reference \cite{jiang2018acquisition}, which means a predicted bounding box with misaligned classification and localization accuracy. This misalignment makes the Non-Maximum Suppression (NMS) procedure unreliable since the NMS only uses the classification score as the metric to rank the proposals, resulting in inaccurate suppression. To alleviate this problem, previous works \cite{tian2019fcos, li2020generalized, jiang2018acquisition, gao2021decoupled, wang2021reconcile, feng2021tood} attempt to make the predictions more harmonious. The route to tackling this issue can be divided into three categories, including reformulating the ranking metric \cite{jiang2018acquisition, gao2021decoupled, tian2019fcos}, focusing on harmonious regions \cite{tian2019fcos, li2020generalized}, and enhancing the dependency between classification and localization tasks \cite{wang2021reconcile, feng2021tood}. IOUNet \cite{jiang2018acquisition} utilizes an extra head to predict the localization-aware score and reformulate the NMS score to pay more attention to the localization task. This paradigm has been popularized by subsequent works such as FCOS \cite{tian2019fcos} and DIR \cite{gao2021decoupled}. In addition, FCOS \cite{tian2019fcos} proposes a centering sampling strategy based on the observation that the center region of GT usually has high classification and regression accuracy. Unlike these methods, GFL \cite{li2020generalized} incorporates the IOU into the classification label. TOOD \cite{feng2021tood} proposes a novel T-head with task alignment learning (TAL) to enhance the interaction between classification and regression tasks. HarmonicDet \cite{wang2021reconcile} improves the prediction consistency from the perspective of loss function excogitation. 

Unlike the previous works, the proposed method generates harmonious predictions from the perspective of knowledge distillation. We first define the Harmony Score (HS) to capture the harmonious relationship between classification score and localization IOU. Then the HS of the teacher model is viewed as prior knowledge and ultimately transferred to the student with the supervision of the proposed HD.
\subsection{Detection-oriented Knowledge Distillation}
Knowledge Distillation (KD) is initially proposed in \cite{hinton2015distilling} to compress cumbersome models and has achieved remarkable progress in image classification. Existing KD-based methods can be broadly divided into response-based methods \cite{hinton2015distilling, li2017learning, yuan2020revisiting, mirzadeh2020improved}, feature-based methods \cite{romero2014fitnets, heo2019knowledge, zagoruyko2016paying}, and relation-based methods \cite{park2019relational, tung2019similarity}. After that, KD has been increasingly applied to the object detection task, achieving significant improvements. Compared with the image classification task, since the image used for object detection normally contains a mass of background pixels, one of the technical routes of the detection-based knowledge distillation is to select suitable distillation regions. Mimicking \cite{li2017mimicking} mimics the feature divergence between teacher and student proposals. FGFI \cite{wang2019distilling} claims that the regions near the ground truths should be regarded as the crucial distillation areas. GID \cite{dai2021general} defines GIs based on the predictions and proposes distillation in an instance-wise manner. DeFeat \cite{Guo_2021_CVPR} confirms that both foreground and background areas are valuable and then proposes a spatial-decoupled distillation to achieve feature imitation. PFI \cite{li2021knowledge} and FRS \cite{zhixing2021distilling} propose a prediction-guided distillation by utilizing the classification score. FGD \cite{yang2022focal} further decouples the feature imitation at the spatial and channel dimensions.

The primary difference between the proposed TFD and the above works is listed as follows. We revisit the selection of valuable feature areas from the perspective of inharmonious task distributions. In particular, both the classification-aware and localization-aware regions are regarded as valuable areas. Moreover, the TWG module is proposed to dynamically assign weights to balance the contributions of these two tasks.
\begin{figure*}[t]
	\centering
	\includegraphics[width=\linewidth]{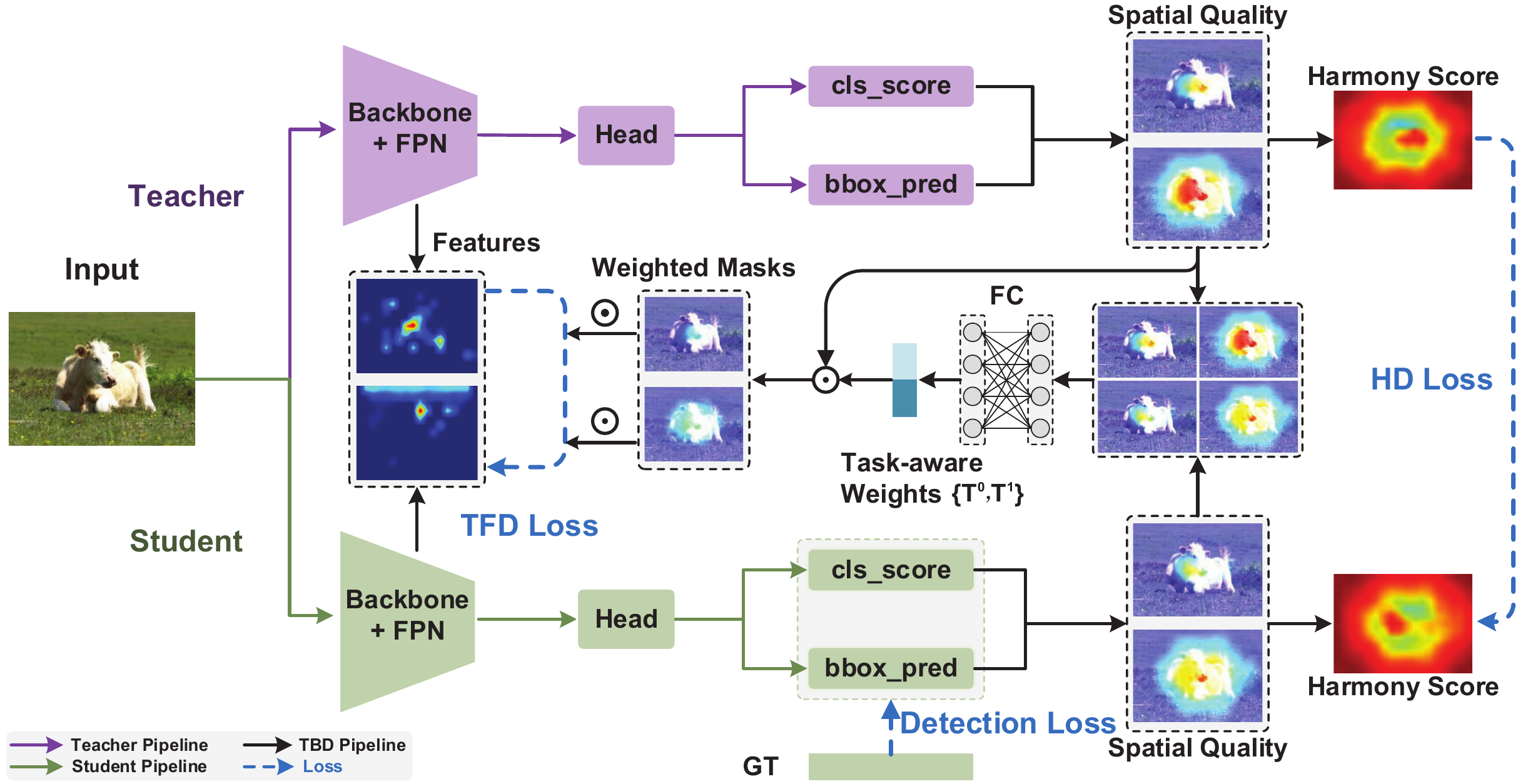}
	\caption{The whole architecture of the proposed TBD. For simplicity, only single-level feature and prediction are shown here.}
	\label{f2}
\end{figure*}
%------------------------------------------------------------------------
\section{Proposed Method}
This section systematically expounds on the overall architecture of the proposed TBD. As demonstrated in Fig.\ref{f2}, the proposed TBD consists of Harmony Distillation (HD) and Task-decoupled Feature Distillation (TFD), elaborated in \ref{HD} and \ref{TFD}, respectively. 
\subsection{Harmony Distillation}\label{HD}
In this subsection, we progressively expound on the derivation of HD. As discussed above, the divergence between classification score and localization quality will lead to incorrect NMS suppression. Therefore, the primary point is to quantify this divergence, called Harmony Score (HS). In general, the derivation of HS consists of two steps: converting the prediction into the classification and localization probabilities and then devising the expression of HS based on the task probabilities.

For each predicted bounding box, $P_{cls}$ and $P_{reg}$ are used to denote the outputs of classification and regression branches, respectively. Concretely, $P_{cls}$ is a score vector of \textit{C} dimensions, where \textit{C} represents the number of classes. Similarly, $P_{reg}$ encodes the regularized offsets from the anchor (anchor box or anchor point) to the actual prediction. 

The defining principle of task-specific probability comprises two parts. First, the probability amplitude should be normalized to $\left [ 0,1 \right ] $. Second, a large probability signifies a precise prediction. For the classification task, the probability $p_{c} $ is graciously generated by succinctly using the normalized maximum activation value:
\begin{equation}\label{eq2}
	p_{c}=\texttt{softmax}\left ( \max_{1\le k\le C} P_{cls}^{k}\right )
\end{equation}
where $P_{cls}^{k}$ is the \textit{k}-th element of classification prediction, and $\texttt{softmax}$ is a spatial-wise softmax function to normalize the reserved classification score.

For the regression task, the implementation is similar to the above one. Concretely, the normalized prediction is firstly converted to the actual bounding box. Then we evaluate the IOU scores among each bounding box and the overall ground truths (GTs). For each predicted box, only the largest IOU score is preserved as the regression probability $p_{r}$.
\begin{equation}\label{eq4}
	p_{r} = \max_{1\le g\le G} IOU\left ( \texttt{decode}(P_{reg}), GT_{g} \right ) 
\end{equation}
where \textit{G} denotes the number of GTs in each image, and $g$ is defined as the corresponding index. $\texttt{decode}$ represents the transformation function to obtain the actual prediction. 

According to Equations \ref{eq2} and \ref{eq4}, given a predicted bounding box, the classification and localization qualities are straightforwardly expressed by a binary tuple $ \left ( p_{c},p_{r}  \right )$. Based on this definition, it is unequivocal to derive the formulation of HS. Theoretically, the definition of HS should satisfy two requirements. First, the formulation is monotonically decreasing. For instance, a diminutive divergence between classification and localization probabilities indicates encouraging consistency; therefore, a high HS should be achieved. Second, the definition should be bounded, which is conducive to avoiding the needless learning dilemma. Based on the above guidelines, HS is arranged through the activation function \texttt{tanh}.
\begin{equation}\label{eq5}
	\Delta p = \left | p_{r} - p_{c}  \right | 
\end{equation}
\begin{equation}\label{eq6}
	HS = 1 - \texttt{tanh}\left ( \Delta p \right ) = 2\times \frac{e^{-\Delta p} }{e^{-\Delta p}+e^{\Delta p}} 
\end{equation}

Since the teacher model performs better than the student in handing inharmonious predictions, a natural thought is to transfer the HS of the teacher as new knowledge to guide the student's learning phase. For distinction, the superscript $t,s$ are used to denote the HS of teacher and student, respectively. In addition, $L1$ Loss is conducted to implement the knowledge transformation. The expression of HD is demonstrated as follows:
\begin{equation}\label{eq7}
	\mathcal{L} _{HD}  =  \sum_{l=1}^{L} \frac{1}{H\times W} \sum_{i=1}^{W} \sum_{j=1}^{H} \left | HS_{i,j,l}^{t}  - HS_{i,j,l}^{s}   \right |  
\end{equation}
where $l$ denotes the $l$-th FPN layer, and $i,j$ are the spatial positions. $W$ and $H$ correspond to the spatial width and height of the prediction. 

Moreover, we notice that Equation \ref{eq7} is calculated with equal contribution of foreground and background predictions. To highlight the contributions of foreground predictions, an IOU-guided harmony loss is established. The $p_{r}^{t}$ generated by the teacher model is employed as a spatial mask to up-weight the significant foreground locations. In addition, a dynamic modulation factor $\sqrt{1+\left | p_{c}^{t}- p_{c}^{s} \right | }$ is introduced to magnify the loss of unfaithful predictions that have large performance gaps with the teacher. Considering the above two points, the spatial mask $\Psi _{i,j,l}$ and Equation \ref{eq7} can be amended as:
\begin{equation}\label{eq8}
	\Psi_{i,j,l} = p_{r,i,j,l}^{t} \times \sqrt{1+\left | p_{c,i,j,l}^{t}- p_{c,i,j,l}^{s} \right | } 
\end{equation}
\begin{equation}\label{eq9}
	\mathcal{L} _{HD}  =\sum_{l=1}^{L} \frac{\sum_{i=1}^{W} \sum_{j=1}^{H} \Psi _{i,j,l}  \left | HS_{i,j,l}^{t}  - HS_{i,j,l}^{s}   \right |}{\sum_{m=1}^{W}\sum_{n=1}^{H}\Psi_{m,n,l} }
\end{equation}

\subsection{Task-decoupled Feature Distillation}\label{TFD}
The proposed Task-decoupled Feature Distillation (TFD) is constructed based on FPN features $F = \left \{ F_{1}, F_{2},..., F_{L}\right \}$. The reasons can be explained on two sides: for one thing, distilling FPN features can facilitate the imitation of both backbone and FPN features. For another, since most recent studies perform distillation on FPN features, it is natural to select FPN features to accomplish a fair comparison. The classic feature mimicking can be denoted as:
\begin{equation}\label{eq10}
	\mathcal{L} _{FPN} = \sum_{l=1}^{L} \sum_{i=1}^{W} \sum_{j=1}^{H}
	(F_{i,j,l}^{t} - \phi (F_{i,j,l}^{s}))^{2}  
\end{equation}
where $\phi\left ( \cdot  \right ) $ denotes the adaptive layer to align the teacher and student features. As can be seen from the definition in Equation \ref{eq10}, the loss will be dominated by the background regions since the background pixels are far more than the foreground ones in the object detection task. Therefore, how to determine the distillation area is a valuable topic. 

Unlike the previous works \cite{yang2022focal, zhixing2021distilling, wang2019distilling, li2021knowledge}, we revisit the selection of crucial areas from the perspective of task-aware spatial distributions. The theoretical reasons for combining classification-aware and localization-aware regions are listed here. On the one hand, the FPN features are the pillar of subsequent classification and regression heads, so combining them is a natural choice. On the other hand, since the previous work \cite{feng2021tood} reveals the inharmonious distributions between two sub-tasks, only applying the classification mask \cite{zhixing2021distilling} might miss some localization-aware regions. Based on the above analysis, the proposed TFD completely utilizes the prediction probabilities $\left ( p_{c}^{t},p_{r}^{t} \right ) $ of the teacher to generate task-aware masks. The mathematics formula is shown as follows:
\begin{align}\label{eq11}	
	\mathcal{L} _{TFD}&= \sum_{l=1}^{L} \frac{\omega _{c}\cdot\sum_{i=1}^{W} \sum_{j=1}^{H} p_{c,i,j,l}^{t}  (F_{i,j,l}^{t} - \phi (F_{i,j,l}^{s}))^{2}}{\sum_{m=1}^{W}\sum_{n=1}^{H}p_{c,m,n,l}^{t} } \notag \\
	&+ \sum_{l=1}^{L} \frac{\omega _{r}\cdot\sum_{i=1}^{W} \sum_{j=1}^{H} p_{r,i,j,l}^{t}  (F_{i,j,l}^{t} - \phi (F_{i,j,l}^{s}))^{2}}{\sum_{m=1}^{W}\sum_{n=1}^{H}p_{r,m,n,l}^{t}  }
\end{align}
where $\omega _{c}$ and $\omega _{r}$ are hyper-parameters to control the weights of classification-aware and localization-aware losses. However, fixed weights applied in Equation \ref{eq11} may suffer from some limitations. For instance, fixed weights are unenviable to adapt to the dynamic inputs comprehensively. In addition, extra hyperparametric optimization overhead is introduced compared with these methods that only utilize the classification mask. Therefore, we propose a Task-collaborative Weight Generation (TWG) module, to dynamically assign weights to overcome these limitations. Motivated by SENet \cite{hu2018squeeze}, TWG only consists of two Fully-Connected (FC) layers and one softmax layer to generate the task-aware weights. Theoretically, the learned weights should be jointly determined by the teacher's prediction and the current learning state of the student. Based on this point, when implementing TWG, the prediction masks $\left ( p_{c}^{t}, p_{r}^{t}, p_{c}^{s}, p_{r}^{s}\right )$ are firstly concatenated at the channel dimension:
\begin{equation}\label{eq13}
	\mathcal{P} = \texttt{concat} \left ( p_{c}^{t}, p_{r}^{t}, p_{c}^{s}, p_{r}^{s} \right ) 
\end{equation}

Then, the concatenated $\mathcal{P}$ is compressed by the average pooling operator. Two lightweight FC layers are subsequently added to generate task-aware weights. Eventually, the softmax function outputs the normalized weights to guarantee that the sum of these weights is 1. Note that we accomplish the implementation of TWG using the most straightforward way to avoid falling into the cumbersome network construction. Consequently, the learned weights can be mathematically expressed as:
\begin{equation}\label{eqw}
	T^{0}, T^{1} = \texttt{softmax}\left ( \texttt{FC}\left ( \texttt{FC}\left ( \texttt{AvgPool}\left ( \mathcal{P } \right ) \right )  \right )  \right )
\end{equation}

Thus, given the task-aware weights $\left \{ T^{0},T^{1} \right \} $, Equation \ref{eq11} can be rewritten as follows:
\begin{align}\label{eq14}
	\mathcal{L} _{TFD-c}&= \sum_{l=1}^{L} \frac{T_{l}^{0}\cdot\sum_{i=1}^{W} \sum_{j=1}^{H} p_{c,i,j,l}^{t}  (F_{i,j,l}^{t} - \phi (F_{i,j,l}^{s}))^{2}}{\sum_{m=1}^{W}\sum_{n=1}^{H}p_{c,m,n,l}^{t}  }    \notag \\
	\mathcal{L} _{TFD-r}&= \sum_{l=1}^{L} \frac{T_{l}^{1} \cdot \sum_{i=1}^{W} \sum_{j=1}^{H} p_{r,i,j,l}^{t}  (F_{i,j,l}^{t} - \phi (F_{i,j,l}^{s}))^{2}}{\sum_{m=1}^{W}\sum_{n=1}^{H}p_{r,m,n,l}^{t}  }   
\end{align}
\begin{equation}\label{eq15}
	\mathcal{L} _{TFD} = \mathcal{L} _{TFD-c} + \mathcal{L} _{TFD-r}
\end{equation}
\subsection{Overall Loss}\label{Loss}
To sum up, the proposed model is trained in an end-to-end manner, and the whole loss includes the original detector loss and the customized distillation loss, demonstrated as follows:
\begin{equation}\label{eq16}
	\mathcal{L} = \mathcal{L} _{detector} + \alpha \cdot \mathcal{L} _{HD} + \beta \cdot \mathcal{L} _{TFD}
\end{equation}
where $\mathcal{L} _{detector}$ is the original detector loss for the student model. $\alpha$ and $\beta$ are hyper-parameters introduced in HD and TFD to balance the distillation loss.
%------------------------------------------------------------------------
\section{Experiments}
\subsection{Datasets and Experimental Settings}
\paragraph{\textbf{Datasets.}}To verify the effectiveness and generalization of the proposed TBD, we conduct abundant experiments on four common datasets, including COCO \cite{lin2014microsoft}, Pascal VOC \cite{everingham2010pascal}, Cityscapes \cite{cordts2016cityscapes} and TJU-DHD \cite{pang2020tju}. Concretely, the main results are firstly reported on COCO, and then other datasets are selected to evaluate the generalization of the proposed method. Following the most common settings for COCO, the distillation models are trained on the 118k training split and evaluated on the \textit{minival} split. In addition,  we also report distillation results trained on COCO \textit{minitrain} split \cite{samet2020houghnet}, which is a curated mini-training set containing about 25k images. For simplicity, we use miniCOCO to describe this dataset. For Pascal VOC, the union of VOC2007 $trainval$ and VOC2012 $trainval$ is chosen for training, whereas the VOC2007 $test$ split is selected for evaluation. For Cityscapes, we use 5000 fine-labeled images for training and testing. For the TJU-DHD dataset, the traffic split is used as the benchmark, which is a diverse high-resolution dataset covering five common categories for generic object detection.
\paragraph{\textbf{Settings.}}All the experiments are conducted based on mmdetection \cite{chen2019mmdetection} toolbox. The hyper-parameters $\alpha, \beta$ in Equation \ref{eq16} are set as $\left \{ \alpha = 5.0, \beta = 0.01 \right \} $. The impact of the hyper-parameters is discussed in Table \ref{tab-hyper}. All the student models are pre-trained on ImageNet \cite{deng2009imagenet}. The inheriting strategy \cite{kang2021instance, yang2022focal} is not applied unless otherwise indicated. In addition, 1x and 2x indicate that the models are trained with 12 and 24 epochs, respectively. The initial learning rate is fixed as 0.01 and 0.02 for one-stage and two-stage detectors, respectively. Moreover, the batch size is set as 16 for all datasets, and the warm-up strategy is applied in the first 500 iterations to make the training procedure more stable. The remaining hyper-parameters are consistent with those in mmdetection. 
\begin{table*}[!t]\centering
	\caption{Comparison results of proposed TBD and existing SOTA methods on COCO \textit{minival}. The symbol - means the results are not available in the original papers. T and S represent the teacher and student models.}	
	\label{t2}
	\resizebox{0.8\textwidth}{!}{
		\begin{tabular}{llccccccc}
			\toprule
			Detector & Model & mAP & $AP_{50}$ & $AP_{75}$ & $AP_{S}$ & $AP_{M}$ & $AP_{L}$ & Reference\\
			\hline
			\multirow{20}{*}{RetinaNet}& T:ResNet101 & 38.9 & 58.0 & 41.5 & 21.0& 42.8&52.4 & ICCV2017  \\
			& S:ResNet50 & 37.4 & 56.7 & 39.6 & 20.0& 40.7&49.7 & ICCV2017  \\
			%\cdashline{2-9}
			&FGFI \cite{wang2019distilling}& 38.6(+1.2) & 58.7 & 41.3 & 21.4& 42.5&51.5 & CVPR2019 \\
			&GID \cite{dai2021general}& 39.1(+1.7)& 59.0 & 42.3 & 22.8& 43.1&52.3 & CVPR2021 \\
			&FRS \cite{zhixing2021distilling} & 39.7(+2.3)& 58.6 & 42.4 & 21.8& 43.5&52.4 & NIPS2021 \\
			&\cite{li2021knowledge} & 39.6(+2.2)& - & - & 21.4& 44.0&52.5 & AAAI2022 \\
			&FGD \cite{yang2022focal}& 39.6(+2.2) & - & - & \textbf{22.9}& 43.7&53.6 & CVPR2022 \\
			&\cellcolor{mycolor}{TBD (ours)}& \textbf{40.0(+2.6)} & \textbf{59.1} & \textbf{42.8} & 22.2& \textbf{44.1}&\textbf{54.0} & NaN \\
			\cline{2-9}
			&T:ResNeXt101& 41.0 & 60.9 & 44.0 & 23.9& 45.2&54.0 & ICCV2017  \\
			&S:ResNet50& 37.4 & 56.7 & 39.6 & 20.0& 40.7&49.7 & ICCV2017 \\
			%\cdashline{2-9}
			&FKD \cite{zhang2020improve} & 39.6(+2.2) & 58.8 & 42.1 & 22.7& 43.3&52.5 & ICLR2021  \\
			&DICOD \cite{guo2021distilling}& 37.9(+0.5) & - & - & 20.5& 41.3&50.5 & NIPS2021  \\
			&FRS \cite{zhixing2021distilling} & 40.1(+2.7) & 59.5 & 42.5 & 21.9& 43.7&54.3 & NIPS2021  \\
			&FGD \cite{yang2022focal} & 40.4(+3.0) & - & - & 23.4& 44.7&54.1 & CVPR2022 \\
			&\cellcolor{mycolor}{TBD (ours)} & \textbf{41.0(+3.6)} & \textbf{60.4} & \textbf{43.8} & \textbf{23.9}& \textbf{45.1}&\textbf{54.7} & NaN  \\
			\cline{2-9}
			&T:ResNet101& 38.9 & 58.0 & 41.5 & 21.0& 42.8&52.4 & ICCV2017  \\
			&S:ResNet18& 33.2 & 51.5 & 35.1 & 17.3& 35.4&44.7 & ICCV2017  \\
			%\cdashline{2-9}
			&FKD \cite{zhang2020improve} & 35.9(+2.7) & 54.4 & 38.0 & 17.9& 39.1&49.4 & ICLR2021  \\
			&FGD \cite{yang2022focal} & 35.9(+2.7) & 53.9 & 38.6 & 18.1& 39.2&49.5 & CVPR2022  \\
			&\cellcolor{mycolor}{TBD (ours)} & \textbf{37.1(+3.9)} & \textbf{55.5} & \textbf{39.8} & \textbf{19.4}& \textbf{40.4}&\textbf{51.6} & NaN  \\
			\hline
			\multirow{11}{*}{Faster RCNN }&T:ResNet101& 39.8 & 60.1 & 43.3 & 22.5& 43.6&52.8 & NIPS2015  \\
			&S:ResNet50& 38.4 & 59.0 & 42.0 & 21.5& 42.1&50.3 & NIPS2015  \\
			%\cdashline{2-9}
			&FGFI \cite{wang2019distilling} & 39.3(+0.9) & 59.8 & 42.9 & 22.5& 42.3&52.2 & CVPR2019  \\
			&GID \cite{dai2021general} & 40.2(+1.8) & 60.7 & 43.8 & 22.7& 44.0&53.2 & CVPR2021  \\
			&FGD \cite{yang2022focal} & 40.4(+2.0) & - & - & 22.8& 44.5&53.5 & CVPR2022  \\
			&\cellcolor{mycolor}{TBD (ours)} & \textbf{40.6(+2.2)} & \textbf{61.0} & \textbf{44.2} & \textbf{23.5}& \textbf{44.7}&\textbf{53.5} & NaN  \\
			\cline{2-9}
			&T:ResNet101& 39.8 & 60.1 & 43.3 & 22.5& 43.6&52.8 & NIPS2015  \\
			&S:ResNet18& 34.5 & 54.6 & 37.2 & 19.2& 36.8&45.2 & NIPS2015  \\
				%\cdashline{2-9}
			&FKD \cite{zhang2020improve} & 37.0(+2.5) & 57.2 & 39.7 & \textbf{19.9}& 39.7&\textbf{50.3} & ICLR2021 \\
			&FGD \cite{yang2022focal} & 37.0(+2.5) & 57.1 & 40.0 & 18.9& 40.6&\textbf{50.3} & CVPR2022 \\
			&\cellcolor{mycolor}{TBD (ours)} & \textbf{37.3(+2.8)} & \textbf{57.3} & \textbf{40.1} & 19.7& \textbf{40.8}& 50.0 & NaN  \\
			\bottomrule
	\end{tabular}}
\end{table*}
\subsection{Main Results}
\subsubsection{Comparison with SOTA methods}
In this part, we compare the proposed TBD with existing state-of-the-art (SOTA) detection-based distillation methods. The main comparison experiments are implemented based on two well-known detectors, including RetinaNet \cite{lin2017focal}and Faster RCNN \cite{ren2015faster}. To verify the superiority of the proposed TBD, seven recent SOTA models \cite{wang2019distilling, zhang2020improve, dai2021general, zhixing2021distilling, li2021knowledge, guo2021distilling, yang2022focal} are used for comparisons, as shown in Table \ref{t2}. In addition, to compare with a recent SOTA, named LD \cite{zheng2022localization}, supplementary experiments are implemented based on GFL \cite{li2020generalized} benchmark. Moreover, referring to the comparison settings of previous works \cite{dai2021general, zhixing2021distilling, yang2022focal}, all the distillation models are conducted with various ResNet/ResNeXt models (\textit{e.g.}, R101-R50, X101-R50). Complementary to the basic configurations, the distillation experiments of using other backbones are shown in Table \ref{t3}.

As presented in Table \ref{t2}, by equipping the proposed TBD, RetinaNet with the ResNet50 student can even exceed the teacher model by a large margin (40.0 vs. 38.9). When an enormous teacher (e.g., ResNeXt101-64x4d) is applied, the lightweight ResNet50 student can still achieve comparable performance. Additionally, according to the results, we can discover that no matter under the guidance of ResNet101 or ResNeXt101, the proposed TBD significantly outperforms the previous SOTA methods. Concretely, with the ResNeXt101 teacher and ResNet50 student, TBD outperforms FRS \cite{zhixing2021distilling} and FGD \cite{yang2022focal} by 0.9 and 0.6, respectively. In addition, when compared with the existing methods on Faster RCNN, consistent gains are achieved, which indicates the effectiveness of the proposed model. Moreover, according to the experimental results shown in Table \ref{t_ld}, the proposed TBD is also compatible with GFL, exceeding the recent LD \cite{zheng2022localization} from 0.6 to 1.7 AP.
\begin{table}[!t]\centering
	\caption{Quantitative results of the proposed TBD and LD \cite{zheng2022localization} on COCO2017 \textit{minival}. The teacher model is ResNet101, and S represents the student model.}	
	\label{t_ld}
	\resizebox{0.48\textwidth}{!}{
		\begin{tabular}{llcccc}
			\toprule
			Detector & Model & mAP & $AP_{S}$ & $AP_{M}$ & $AP_{L}$\\
			\hline
			\multirow{9}{*}{GFL}&S:ResNet18& 35.7 & 19.4& 38.8&47.5 \\
			%\cdashline{2-8}
			&LD \cite{zheng2022localization} & 37.5(+1.8) & 20.2& 41.2&49.4 \\
			&TBD (ours) & \textbf{39.2(+3.5)} & \textbf{22.5}& \textbf{43.0}&\textbf{51.9}  \\
			\cline{2-6}
			&S:ResNet34& 38.9 & 21.5& 42.8&51.4 \\
			%\cdashline{2-8}
			&LD \cite{zheng2022localization} & 41.0(+2.1) & 23.2& 45.0&54.2 \\
			&TBD (ours) & \textbf{41.6(+2.7)} & \textbf{24.4}& \textbf{45.7}&\textbf{54.2}  \\
			\cline{2-6}
			&S:ResNet50& 40.2 & 23.3& 44.0&52.2 \\
			%\cdashline{2-8}
			&LD \cite{zheng2022localization} & 42.1(+1.9) & 24.5& 46.2&54.8 \\
			&TBD (ours) & \textbf{43.4(+3.2)} & \textbf{25.9}& \textbf{47.6}&\textbf{55.6}  \\
			\bottomrule
	\end{tabular}}
\end{table}
\begin{table}[t]\centering
	\caption{Results of applying TBD among diverse backbones on COCO. All the experiments are based on RetinaNet with the 1x training schedule.}
	\label{t3}
	\resizebox{0.48\textwidth}{!}{
	\begin{tabular}{lllc}
		\toprule
		Student & Teacher & Distillation Type & mAP  \\
		\hline
		\multirow{5}{*}{ResNet18}& - & Baseline & 31.9  \\
		& ResNet50 & CNN-CNN  & 34.7 (+2.8)  \\
		& ResNet101 & CNN-CNN  & 35.3 (+3.4) \\
		& PVTb0 \cite{wang2022pvt} & Trans-CNN  & 35.1 (+3.2) \\
		& PVTb1 \cite{wang2022pvt} & Trans-CNN  & 36.2 (+4.3) \\
		\hline
		\multirow{3}{*}{ResNet50}& - & Baseline &36.5 \\
		& PVTb1 \cite{wang2022pvt} & Trans-CNN &39.5 (+3.0)\\
		& PVTb2 \cite{wang2022pvt} & Trans-CNN &40.5 (+4.0)\\
		\hline
		\multirow{2}{*}{\makecell[l]{RegNetX \\ 800MF\\ \cite{radosavovic2020designing}}}& - & Baseline &35.6 \\
		& \makecell[l]{RegNetX \\ 3.2GF\cite{radosavovic2020designing}} & CNN-CNN &38.2 (+2.6)\\
		\hline
		\multirow{2}{*}{PVTb0}& - & Baseline &37.1 \\
		& PVTb2 & Trans-Trans & 39.7 (+2.6) \\
		\bottomrule
	\end{tabular}}
\end{table}
\subsubsection{Results of TBD using various teacher-student configurations}\label{4-2-2}
This part shows the results of using diverse teacher-student configurations on COCO, including CNN-CNN, Transformer-Transformer, and Transformer-CNN. All the experiments are based on the RetinaNet. The correlative results are presented in Table \ref{t3}. Overall, consistent proceeds are obtained, declaring that the proposed TBD is feasible, efficient, and stable. Specifically, using CNN teachers, the ResNet18 student with TBD can effectively obtain performance gains from 2.8 to 4.3. When the heterogeneous teacher-student pair is applied (\textit{e.g.,} from PVT \cite{wang2022pvt} to ResNet), the ResNet18 student overcomes the architecture divergence between teacher and student and finally achieves 4.3 mAP gains. Moreover, the improvements of experiments based on PVT \cite{wang2022pvt} and RegNet \cite{radosavovic2020designing} verify the superiority of the proposed method, as well.
\begin{table}[t]\centering
	\caption{Results of applying TBD on diverse detectors based on COCO. }
	\label{t4}
	\resizebox{0.48\textwidth}{!}{
	\begin{tabular}{lccccc}
		\toprule
		Detector & Distill & mAP & $AP_{s}$ & $AP_{m}$ & $AP_{l}$   \\
		\hline
		\multirow{2}{*}{Faster RCNN }& $\times$ & 33.2 & 18.2 & 35.9 & 43.2 \\
		& $\surd$ & 35.4 (+2.2)  & 19.6 & 38.8 & 46.4  \\
		\hline
		\multirow{2}{*}{Dynamic RCNN }& $\times$ & 34.9 & 18.3 & 37.2 & 47.7  \\
		& $\surd$ & 36.9 (+2.0) & 19.6 & 39.7 & 49.4  \\
		\hline
		\multirow{2}{*}{RetinaNet}& $\times$ & 31.9 & 16.4 & 34.6 & 43.4  \\
		& $\surd$ & 34.7 (+2.8) & 17.9 & 38.0 & 47.6  \\
		\hline
		\multirow{2}{*}{GFL }& $\times$ & 35.7 & 19.4 & 38.8 & 47.5  \\
		& $\surd$ & 38.2 (+2.5)& 20.6 & 41.7 & 50.2  \\
		\hline
		\multirow{2}{*}{FSAF}& $\times$ & 32.4 & 17.1 & 35.5 & 42.3   \\
		& $\surd$ & 35.1 (+2.7) & 17.1 & 38.1 & 47.3 \\
		\hline
		\multirow{2}{*}{FreeAnchor }& $\times$ & 34.0 & 18.1 & 36.3 & 46.5  \\
		& $\surd$ & 37.2 (+3.2) & 19.2 & 40.2 & 50.5  \\
		\bottomrule
	\end{tabular}}
\end{table}
\subsubsection{Results of TBD on various detectors}\label{4-2-3}
In this piece, we implement the proposed TBD on six prevalent detectors, including two-stage detector Faster RCNN \cite{ren2015faster}, Dynamic RCNN \cite{zhang2020dynamic}, and one-stage detector FreeAnchor \cite{zhang2019freeanchor}, RetinaNet \cite{lin2017focal}, GFL \cite{li2020generalized} and FSAF \cite{zhu2019feature}. Here ResNet50 is adopted as the teacher while the lightweight ResNet18 is set as the student. All the models are trained with the 1x learning paradigm. When the proposed TBD is adaptive to the two-stage detector, the classification and localization masks are generated on Region Proposal Network (RPN). Table \ref{t4} summarizes the detailed results. Overall, the consistent improvements indicate that the proposed TBD is compatible with mainstream detectors.
\begin{table}[t]\centering
	\caption{Results of applying TBD on other datasets. The last row of data in miniCOCO is obtained using the teacher model trained on full COCO split.}
	\label{t5}
	\resizebox{0.48\textwidth}{!}{
	\begin{tabular}{lccccc}
		\toprule
		Datasets & Distill & mAP & $AP_{s}$ & $AP_{m}$ & $AP_{l}$   \\
		\hline
		\multirow{3}{*}{miniCOCO \cite{samet2020houghnet}}& $\times$ & 19.8 & 9.2 & 21.4 & 26.9  \\
		& $\surd$ & 25.5 (+5.7)  & 12.2 & 27.5 & 34.2  \\
		& $\surd$ & 29.0 (+9.2) &  13.1 & 31.4 & 39.9 \\
		\hline
		\multirow{2}{*}{Pascal VOC \cite{everingham2010pascal}}& $\times$ & 48.8 & 17.5 & 32.0 & 54.6  \\
		& $\surd$ & 52.7 (+3.9) & 18.6 & 35.9 & 58.4  \\
		\hline
		\multirow{2}{*}{Cityscapes \cite{cordts2016cityscapes}}& $\times$ & 30.1 & 10.6 & 31.6 & 47.7  \\
		& $\surd$ & 33.9 (+3.8) & 12.5 & 34.5 & 54.6 \\
		\hline
		\multirow{2}{*}{TJU-DHD \cite{pang2020tju}}& $\times$ & 50.4 & 19.2 & 47.1 & 65.9   \\
		& $\surd$ & 52.5 (+2.1) & 21.7 & 48.8 & 68.7 \\
		\bottomrule
	\end{tabular}}
\end{table}
\subsubsection{Results of TBD on other datasets}\label{4-2-4}
The above experiments are completely implemented based on MS COCO. In this piece, we evaluate our TBD on other datasets. Concretely, the widely used Pascal VOC, miniCOCO, TJU-DHD, and Cityscapes are introduced to evaluate the performance of TBD on small datasets. Analogously, all the models are trained with RetinaNet-R18 baseline with ResNet50 as the teacher. The complete results are shown in Table \ref{t5}. For small datasets, we notice that remarkable improvements are achieved with the assistance of the proposed TBD. For example, TBD dramatically improves the vanilla student model by 5.7 mAP on miniCOCO. The progress can be extended to 9.2 using the teacher training on the complete COCO train split. Furthermore, the consistent gains in Table \ref{t5} manifest that the proposed TBD performs magnificently among multifarious datasets.
\subsection{Ablation Study}
In this part, we conduct abundant experiments based on RetinaNet to demonstrate the effectiveness of each component and explore some implementation details of TBD. The analytical experiments are implemented on the ResNet18 with ResNet50 as the teacher. The whole experiments are trained with the 1x learning schedule. 
\subsubsection{Ablation study of each component}
As presented in Table \ref{tab:tab2}, the vanilla student model achieves 31.9 mAP. When the proposed method is applied, both HD and TFD can consistently promote student performance. Concretely, HD obtains 1.3 gains while TFD harvests 2.5 improvements. In addition, the combination of HD and TFD brings the maximum promotion (\textit{i.e.}, 2.8 mAP).
\begin{table}[t]\centering
	\caption{The individual results of HD and TFD on COCO.}
	\label{tab:tab2}
	\resizebox{0.48\textwidth}{!}{
	\begin{tabular}{ccccccc}
		\toprule
		Model & HD & TFD & mAP & $AP_{S}$ & $AP_{M}$ & $AP_{L}$ \\
		\hline
		T: R50 & - & - & 36.5 & 20.4 & 40.3 & 48.1 \\
		\hline
		\multirow{4}{*}{S: R18} & - & - & 31.9 & 16.4 & 34.6 & 43.4 \\
		& $\surd $ &  & 33.2 (+1.3)& 17.2 & 36.2 & 44.2 \\
		&  & $\surd $ & 34.4 (+2.5) & 17.6 & 37.8 & 46.7 \\
		& $\surd $ & $\surd $ & \textbf{34.7 (+2.8)} & \textbf{17.9} &\textbf{ 38.0} &\textbf{ 47.6} \\
		\bottomrule
	\end{tabular}}
\end{table}
\begin{table}[t]\centering
	\caption{The comparison of disparate definitions of HS. The experiments are conducted with RetinaNet on COCO2017. The teacher model is ResNet50, while the student is ResNet18. }
	\label{tab:tab3}
	\resizebox{0.48\textwidth}{!}{
	\begin{tabular}{lllcccc}
		\toprule
		 HS & Loss  & mAP & $AP_{S}$ & $AP_{M}$ & $AP_{L}$ \\
		\hline
		- & - & 31.9 & 16.4 & 34.6 & 43.4 \\
		\hline
		$HS_{exp}$ & L2 & 33.0 (+1.1)& 16.8 & 35.7 & 44.6 \\
		$HS_{exp}$ & L1 & 33.0 (+1.1) & 16.8 & 35.8 & 44.3 \\
		$HS_{log}$ & L2 & 32.8 (+0.9) & 16.9 &35.5 & 44.2 \\
		$HS_{log}$ & L1 & \textbf{33.3 (+1.4)} & 16.9 &36.2 & \textbf{45.2} \\
		$HS_{tanh}$ & L2 & 33.0 (+1.1) & 17.0 &36.0 & 44.4 \\
		$HS_{tanh}$ & L1 & 33.2 (+1.3) & \textbf{17.2} &\textbf{36.2} & 44.2 \\
		\bottomrule
	\end{tabular}}
\end{table}
\subsubsection{Ablation study of HD}
\paragraph{\textbf{Definition of HS.}} In this piece, we dive deeper into the definition of HS. Two variants of HS are customized, named $HS_{exp}$ and $HS_{log}$, respectively. In addition, L1 and L2 Loss are also introduced to evaluate the design of the distillation loss function. The expressions of $HS_{exp}$ and $HS_{log}$ are listed as follows.
\begin{align}\label{eq17}
	HS_{exp} &=  e^{-\left | p_{c} - p_{r} \right | }  \notag \\
	HS_{log} &= \frac{1}{log(e + \left | p_{c} - p_{r}   \right | )} 
\end{align}

To make it convenient to distinguish, we use $HS_{tanh}$ to represent the HS definition in Equation \ref{eq6}. The overall experiment results are shown in Table \ref{tab:tab3}. We can discover that all the implementations of HS can consistently boost the baseline performance. In addition,  $HS_{log}$ and $HS_{tanh}$ achieve superior performance compared with other definitions, which can improve the student model by 1.4 and 1.3. Moreover, we notice that $HS_{tanh}$ slightly outperforms $HS_{log}$ when HD is combined with TFD, so we prefer $HS_{tanh}$ as the ultimate representation.
\begin{table}[t]\centering
	\caption{Comparisons between HD and other response-based distillation methods. \texttt{cls} and \texttt{loc} means the classification and localization task.}
	\label{tab:tab4}
	\resizebox{0.48\textwidth}{!}{
		\begin{tabular}{llcc}
			\toprule
			Model & Distillation  & mAP & Knowledge \\
			\hline
			\multirow{6}{*}{\makecell[c]{T: R50 \\ S: R18}} & Baseline  & 31.9 & None\\
			& KD \cite{hinton2015distilling} & 32.4 (+0.5)& \texttt{cls} logits\\
			& FRS \cite{zhixing2021distilling} & 33.0 (+1.1)& \texttt{cls} logits\\
			& RM \cite{li2021knowledge} & 33.3 (+1.4) & Anchor rank\\
			& HD& 33.2 (+1.3) & \multirow{2}{*}{\makecell[c]{Relationship between \\ \texttt{cls} and \texttt{loc}}}  \\
			& HD$\dagger$ & \textbf{33.9 (+2.0)} \\
			\bottomrule
	\end{tabular}}
	\begin{tablenotes}
		\footnotesize
		\item $\dagger$ means we retrain the proposed HD with the configuration of RM. The values of KD and FRS are our reproduction results. 
	\end{tablenotes}
\end{table}
\paragraph{\textbf{Relationship with other response-based methods.}}Table \ref{tab:tab4} compares the proposed HD with other response-based distillation methods such as KD \cite{hinton2015distilling}, FRS \cite{zhixing2021distilling}, and RM \cite{li2021knowledge}. Although the conventional soft label distillation \cite{hinton2015distilling} obtains remarkable improvements in the image classification task, the promotion is unsatisfactory when applied to the object detection task. FRS \cite{zhixing2021distilling} ameliorates the traditional KD by introducing a feature richness mask. Unlike these methods using classification logits, the proposed HD captures the relationship between classification and localization tasks as the prior knowledge and outperforms KD and FRS by 0.8 and 0.2. When compared with RM, our proposed HD$\dagger$ also shows distinct advantages, demonstrating the tremendous potential of the proposed HD. 

\begin{table}[t]\centering
	\caption{Quantitative results of the proposed TFD. \textit{cls} and \textit{reg} represent using the classification-aware and regression-aware masks, respectively. \textit{fixed} means the weights are optimized as the fixed hyper-parameters. In contrast, \textit{dynamic} denotes the weights are generated by the proposed TWG.}
	\label{tab:tab5}
	\resizebox{0.48\textwidth}{!}{
	\begin{tabular}{cccccc}
		\toprule
		Mask & Weight & mAP & $AP_{S}$ & $AP_{M}$ & $AP_{L}$ \\
		\hline
		- & - & 31.9 & 16.4 & 34.6 & 43.4 \\
		\textit{whole} & \textit{fixed} &33.3 (+1.4)& 17.6 & 36.3 & 44.7 \\
		\textit{cls}  & \textit{fixed}&34.0 (+2.1)& 17.4 & 37.4 & 46.7 \\
		\textit{reg}   &\textit{fixed} &34.0 (+2.1)& 17.4 & 37.5 & 46.5 \\
		\textit{cls + reg}& \textit{fixed}&34.1 (+2.2) & 17.3 & 37.6 & 46.4 \\
		\textit{cls + reg}& \textit{dynamic} &\textbf{34.4 (+2.5)} & \textbf{17.6} &\textbf{37.8} & \textbf{46.7} \\
		\bottomrule
	\end{tabular}}
\end{table}
\begin{table}[t]\centering
	\caption{Comparison results of the proposed TFD and other feature imitation methods.}
	\label{tab:tab6}
	\resizebox{0.45\textwidth}{!}{
		\begin{tabular}{llcc}
			\toprule
			Model & Distillation  & mAP  & Key Region\\
			\hline
			\multirow{6}{*}{\makecell[c]{T: R50 \\ S: R18}} & Baseline  & 31.9 & \textit{cls}\\
			& FitNet \cite{romero2014fitnets} & 33.3 (+1.4)& \textit{cls}\\
			& FRS \cite{zhixing2021distilling} & 34.0 (+2.1)& \textit{cls}\\
			& PFI \cite{li2021knowledge} & 34.2 (+2.3) & \textit{cls}\\
			& TFD (ours)& 34.4 (+2.5) & \textit{cls} + \textit{loc}\\
			& TFD$\dagger$ (ours)& \textbf{35.1 (+3.2)}& \textit{cls} + \textit{loc} \\
			\bottomrule
	\end{tabular}}
	\begin{tablenotes}
		\footnotesize
		\item $\dagger$ means we retrain the proposed TFD with the configuration of PFI. The values of FitNet and FRS are our reproduction results.  
	\end{tablenotes}
\end{table} 
\begin{figure*}[t]
	\centering
	\includegraphics[width=0.9\linewidth]{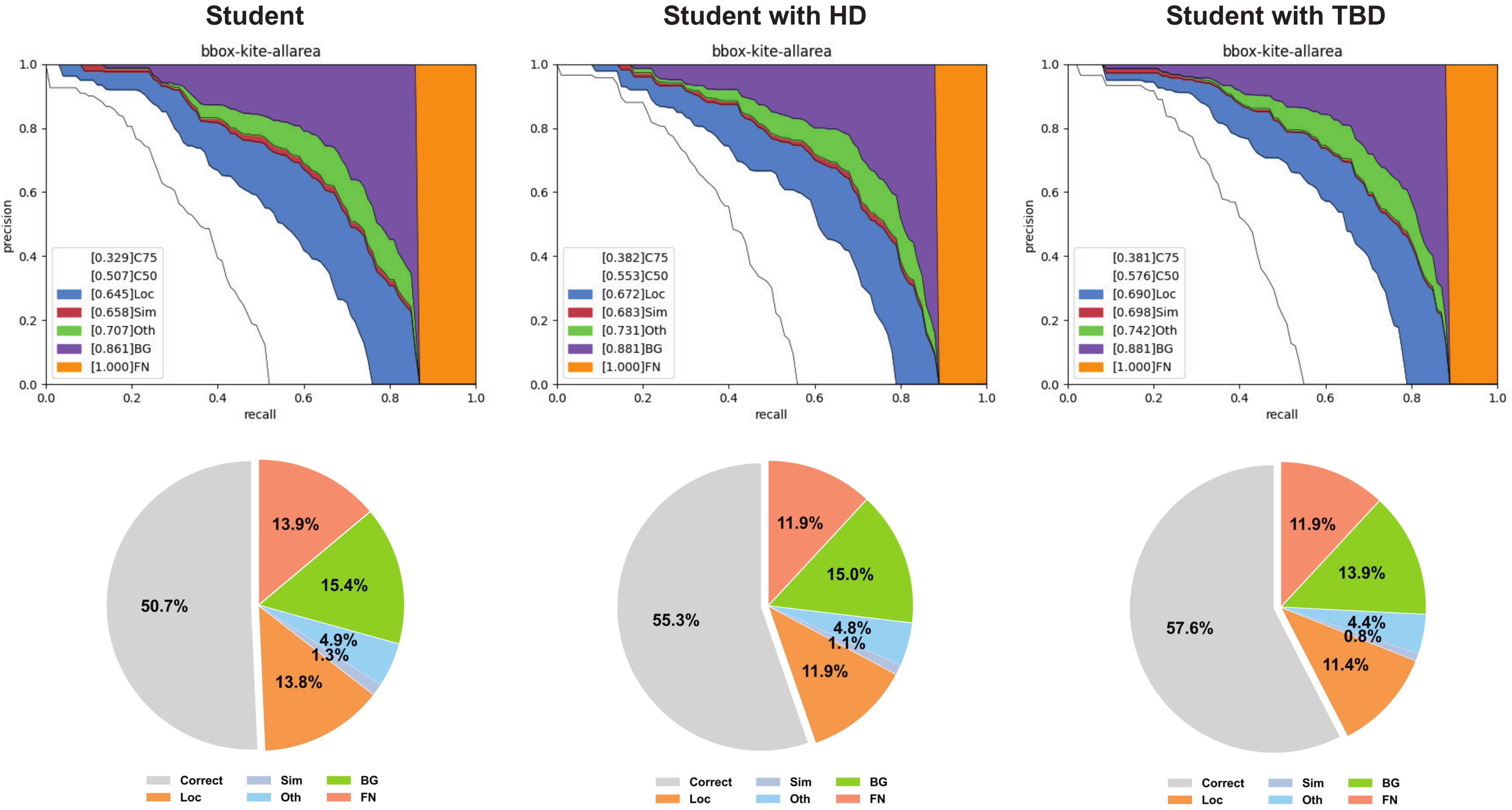}
	\caption{PR curves and error analyses among diverse models. 'Correct': predictions with correct label and $IOU>0.5$. 'Loc': predictions with correct label but $0.1<IOU <0.5$. 'Sim': predictions with an incorrect label but accurate supercategory. 'Oth': predictions with an incorrect label. "BG": false positives predicted on background regions. 'FN': false negatives. }
	\label{f2_app}
\end{figure*}
\begin{table*}[t]\centering
	\caption{The impact of applying different $\alpha$ and $\beta$.}
	\label{tab-hyper}
	\resizebox{0.72\textwidth}{!}{
		\begin{tabular}{ccccc|ccccc}
			\toprule
			$\alpha$& 10.0 & 7.5 & 5.0 & 2.5 & $\beta$& 0.015 & 0.01 & 0.005 & 0.0025\\ \hline
			mAP& 34.6 & 34.6  & \textbf{34.7}  & 34.6 & mAP &34.6& \textbf{34.7}  & 34.5 & 34.5  \\
			\bottomrule
	\end{tabular}}
\end{table*}
\subsubsection{Ablation study of TFD} 
\paragraph{\textbf{Impact of decoupling task-aware masks.}}This part verifies the effectiveness of decoupling classification-aware and localization-aware regions. Concretely, we delicately excogitate several comparison experiments for quantitative verification. As shown in Table \ref{tab:tab5}, compared with distilling on the whole feature map, utilizing both the classification mask $p_{c}^{t}$ and localization mask $p_{r}^{t}$ can obviously promote effective distillation. Technically, $p_{c}^{t}$ or $p_{r}^{t}$ only capture the corresponding task information, which might neglect potential clues about another one since the spatial distributions of $p_{c}^{t}$ and $p_{r}^{t}$ might be disparate. In addition, we observe that the key to integrating the classification-aware and localization-aware regions is how to balance the contribution of each task. Obviously, the information on input characteristics and current training status cannot be considered comprehensively by using a fixed weight scheme, leading to inconspicuous improvement. When equipped with the proposed TWG, the performance of TFD obtains a noticeable promotion (\textit{i.e.}, 34.4 vs. 34.0).
\paragraph{\textbf{Relationship with other feature-based methods.}} Similarly, we compare the proposed TFD with FitNet \cite{romero2014fitnets} and other prediction-guided feature imitations \cite{li2021knowledge, zhixing2021distilling}, and the overall results are presented in Table \ref{tab:tab6}. The proposed TBD surpasses the FitNet by a large margin (1.1 AP). Compared with the recent SOTA models such as FRS and PFI, the proposed TFD still shows its superiority. In particular, FRS and PFI only utilize the information of the classification branch to generate the feature mask, thus resulting in suboptimal results (34.0 and 34.2). On the contrary, the proposed TFD combines the superiority of classification and regression tasks and consistently outperforms them by 0.4 and 0.9.
\subsubsection{Ablation study of hyper-parameters} 
Compared with other detection-based KD methods such as FGD \cite{yang2022focal}, the number of hyper-parameters introduced in this paper is significantly reduced (2 vs. 5) so that it is not difficult to prune them. Concretely, the hyper-parameter analysis is conducted based on RetinaNet-R18 student with knowledge distilling from RetinaNet-R50 teacher. The detailed  comparison results of various values of $\alpha, \beta$ are shown in Table \ref{tab-hyper}. Obviously, the performance of the proposed TBD is sightly affected by these two hyper-parameters with only 0.2 mAP fluctuation. Therefore, $\alpha=5.0, \beta=0.01$ are set as the default configurations. 
\begin{table}[t]\centering
	\caption{Comparison results of the proposed TBD and related detection methods. The experiments are constructed on the detectors with ResNet50. Besides, the detectors with ResNeXt101 is served as the teacher model. $\dagger\dagger$ means the inhering strategy \cite{yang2022focal} is applied to help the convergence of the student model.}
	\label{tab:tab7}
	\resizebox{0.4\textwidth}{!}{
	\begin{tabular}{llc}
		\toprule
		Baseline & Method  & mAP  \\
		\hline
		\multirow{3}{*}{\makecell[l]{Faster RCNN \\ ResNet50}} & HarmonicDet \cite{wang2021reconcile} & 39.2 \\
		& TBD (ours)& 40.3   \\
		& TBD$\dagger\dagger$ (ours)& \textbf{40.6}  \\
		\hline
		\multirow{3}{*}{\makecell[l]{RetinaNet \\ ResNet50}} & HarmonicDet \cite{wang2021reconcile} & 37.6   \\
		& TBD (ours)& 40.0  \\
		& TBD$\dagger\dagger$ (ours)& \textbf{40.3} \\
		\bottomrule
	\end{tabular}}
\end{table}
\subsubsection{Comparison between TBD and related detection-based methods}
As mentioned above, the generation of harmonious predictions can be promoted by a series of methods based on the detector itself, such as devising a customized training strategy \cite{wang2021reconcile}. In this part, we compared our TBD with the recent solution named HarmonicDet \cite{wang2021reconcile} to show the superiority of the KD-based method. The results are summarized in Table \ref{tab:tab7}. According to the results, the proposed TBD outperforms HarmonicDet by 1.1 and 2.4 on Faster RCNN and RetinaNet, indicating the impressive potential of applying knowledge distillation to alleviate the inherent detection problem. In addition, after using the initialization strategy proposed in \cite{kang2021instance}, the proposed TBD allows for faster convergence and achieves more satisfactory performance.
\subsection{Analysis and Visualization}
\subsubsection{Error Analysis}
We use the official COCO toolbox \cite{lin2014microsoft} to conduct error analysis between RetinaNet-R18 and RetinaNet-R18 with the proposed TBD. Note that the RetinaNet-R50 is chosen as the cumbersome teacher. As presented in Fig. \ref{f2_app}, after equipping with the proposed HD, the localization error (Loc) is significantly decreased (\textit{i.e.}, from 13.8 to 11.9). When incorporated with TFD, both the localization and classification errors are further declined. Ultimately, the proportion of correct predictions is boosted from 50.7 to 57.6, verifying the effectiveness of the proposed TBD. 
\subsubsection{Visualization}
\begin{figure}[t!]
	\centering
	\begin{subfigure}[t]{0.45\textwidth}
		\centering
		\includegraphics[width=\textwidth]{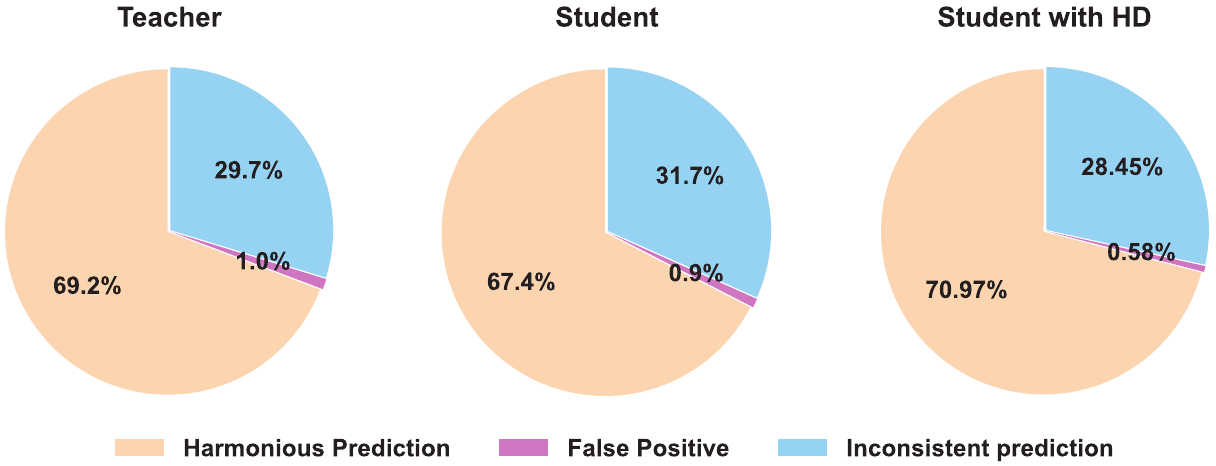}
		\caption{}
		\label{f_iou:a}
	\end{subfigure}
	\begin{subfigure}[t]{0.45\textwidth}
		\centering
		\includegraphics[width=\textwidth]{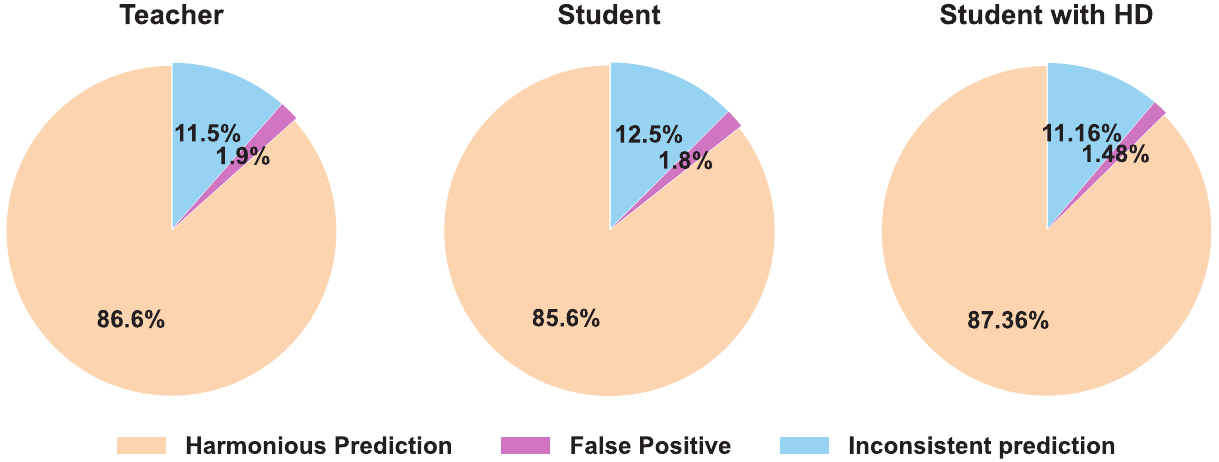}
		\caption{}
		\label{f_iou:b}
	\end{subfigure}
	\caption{The proportions of harmonious and inconsistent predictions between vanilla student and student with proposed HD. The predictions with the score larger than 0.9 and 0.8 are counted in (a) and (b), respectively.}
	\label{f_iou}
\end{figure}
\begin{figure}[t]
	\centering
	\includegraphics[width=0.95\linewidth]{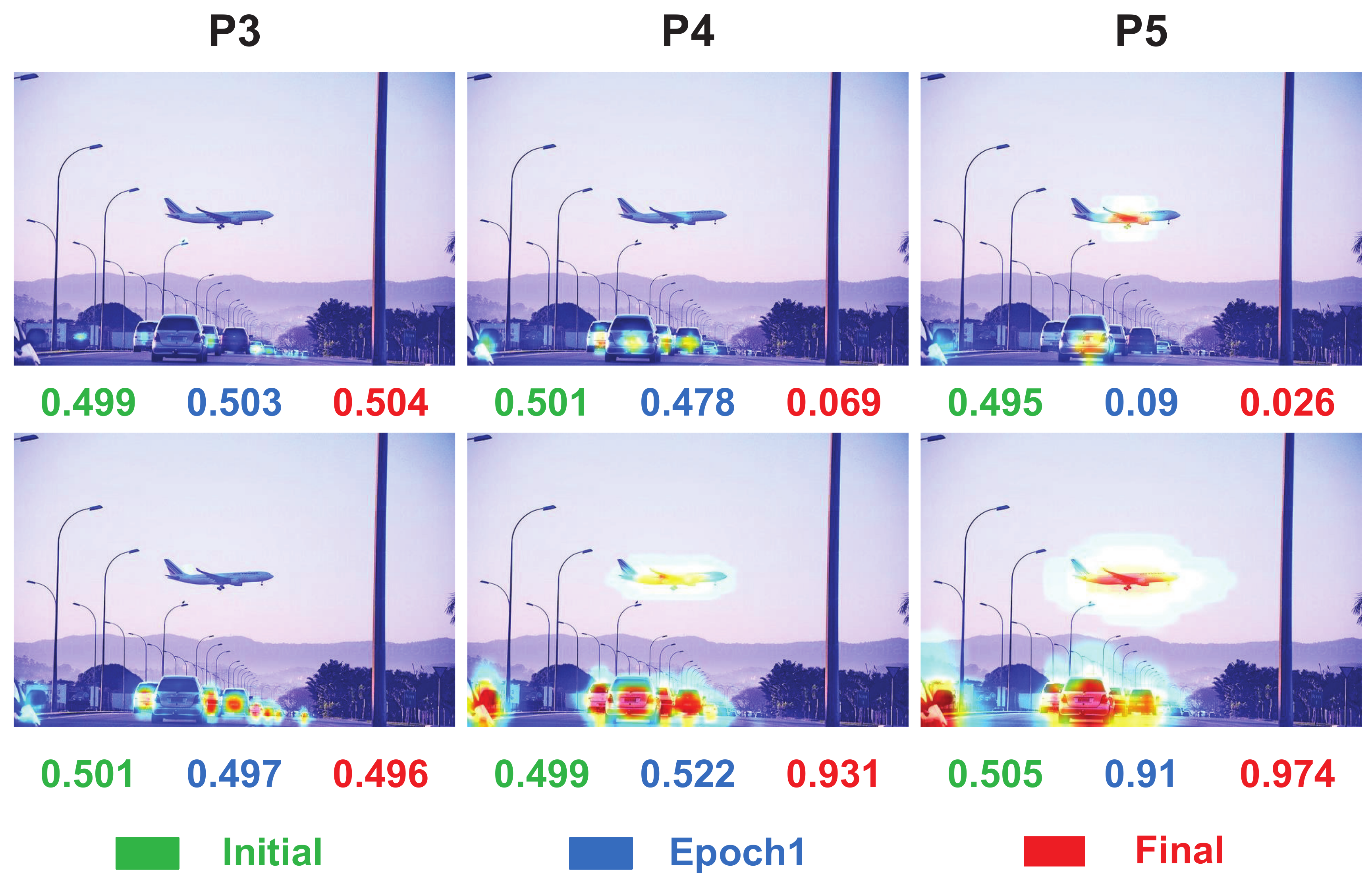}
	\caption{Visualization of the classification-aware and localization-aware masks with the corresponding learned weights. The top row sub-figures are the classification masks, and the bottom are the regression masks.}
	\label{f_mask}
\end{figure}
\begin{figure}[t]
	\centering
	\includegraphics[width=\linewidth]{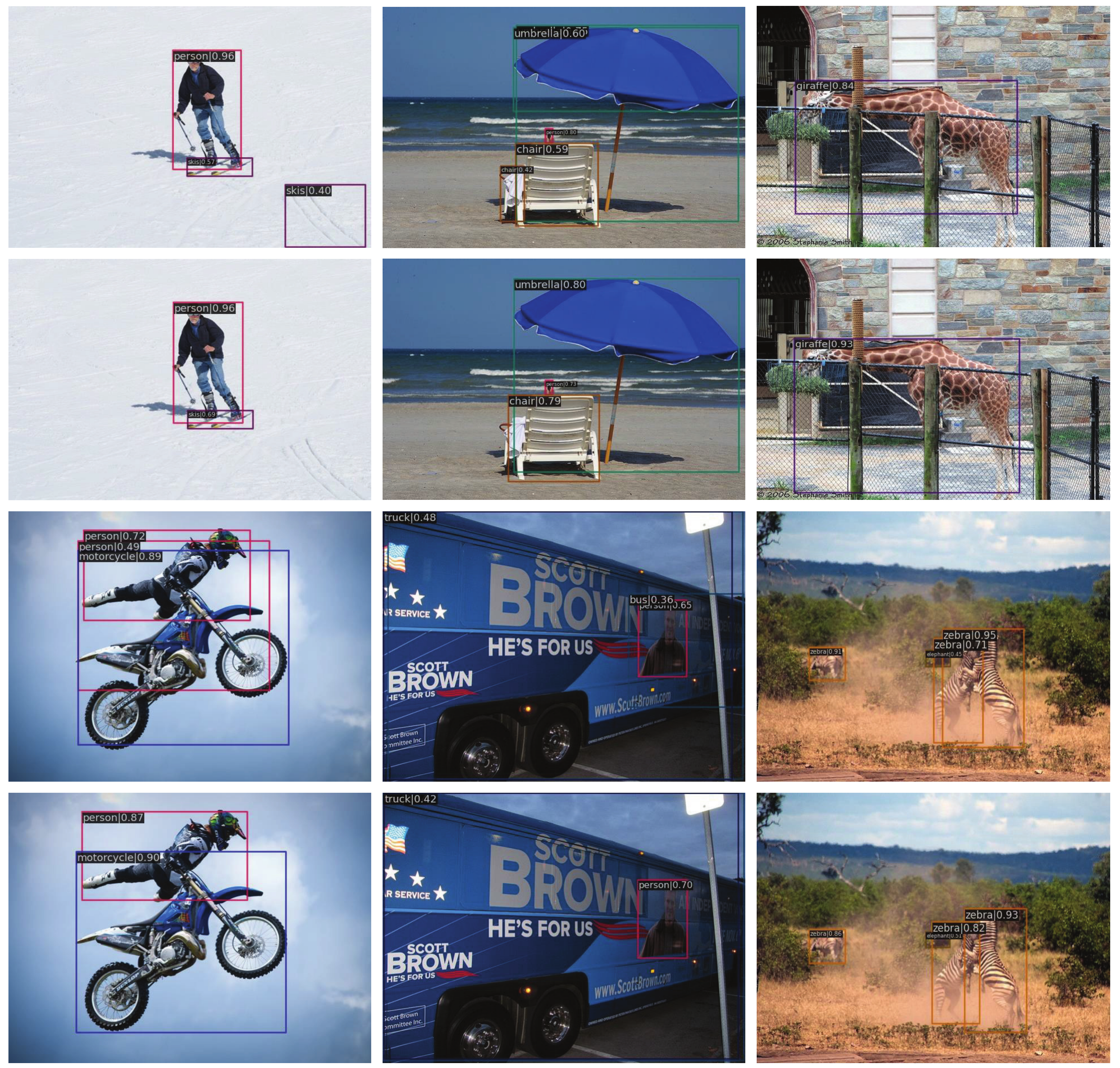}
	\caption{Qualitative comparisons between vanilla RetinaNet-R50 (the top sub-figures) and RetinaNet-R50 with the proposed TBD (the bottom sub-figures).}
	\label{f_det_results}
	%\Description{A woman and a girl in white dresses sit in an open car.}
\end{figure}

\paragraph{\textbf{Visualization of Harmony Distillation.}}We compare the proportions of harmonious predictions of easy-classified samples between the original student model and the model with the proposed HD. The corresponding results of the teacher model are also shown here for reference. The overall results of these models are depicted in Fig. \ref{f_iou}, where the predictions with classification scores larger than 0.9 and 0.8 are selected, respectively. As can be seen from the left and middle sub-figures of Fig. \ref{f_iou:a} and \ref{f_iou:b}, the teacher model is more inclined to generate high-quality predictions (69.2 vs. 67.4, 86.6 vs. 85.6), illustrating that the teacher model has the capacity to transfer knowledge to the lightweight student. Furthermore, the student model's proportions of harmonious predictions tremendously increased from 67.4 to 70.97 and 85.6 to 87.36, even exceeding the teacher model. Besides, we also observe that the amount of False Positives (FPs) is partly reduced. We explain that some FPs with IOU closer to 0.5 can be turned into True Positives (TPs) with the help of HD.  
\paragraph{\textbf{Visualization of task-ware masks and the learned weights.}} In this part, we provide several visualization results of the proposed TFD. Concretely, we visualize the task-aware masks and the learned weights in Fig. \ref{f_mask}. Two meaningful observations can be discovered. For one thing, the distributions of classification and localization masks are disparate. The classification task concentrates on significant parts of the instance, while the regression task encodes rich information between foreground and background. For another, the learned weights behave diversely at different FPN levels and training stages. The task-aware weights tend to be evenly distributed in the initial phase with random initialization. Owing to the proposed TWG, these weights are rapidly modulated by the teacher's predictions and the student's current learning state. 
\paragraph{\textbf{Visualization of detection results.}} Qualitative comparisons between the vanilla student and student with TBD are demonstrated in Fig. \ref{f_det_results}. Compared with the vanilla student, the proposed TBD achieves more credible predictions, such as accurate bounding boxes and fewer duplicates, indicating the effectiveness of our method.
\section{Conclusion}
This paper thoroughly investigates the impact of the inharmonious distributions between classification and regression tasks on distilling object detectors. To alleviate this limitation, we propose a novel Task-Balanced Distillation (TBD), composed of Harmony Distillation (HD) and Task-decoupled Distillation (TFD). HD enhances the harmonious predictions for the student by aligning the Harmony Score (HS) between the teacher and student to make the NMS more credible. In addition, TFD dynamically combines the classification-aware and localization-aware regions as the meaningful regions for distilling features. Extensive experiments among various datasets and detectors verify the effectiveness and generalization of the proposed method.
%-------------------------------------------------------------------------

%%%%%%%%% REFERENCES
{\small
\bibliographystyle{ieee_fullname}

}

\end{document}